\documentclass[letterpaper, 10 pt, conference]{ieeeconf}
\IEEEoverridecommandlockouts                              

\usepackage{cite}

\usepackage{amsmath,amssymb,amsfonts}
\usepackage{algorithmic}
\usepackage{graphicx}
\graphicspath{ {./figures/} }
\usepackage{textcomp}
\usepackage{xcolor}
\usepackage{bbm}
\usepackage{comment}
\usepackage{paralist}
\usepackage[caption=false]{subfig}
\usepackage{scalerel}    

\usepackage{amsthm}
\theoremstyle{plain}

\usepackage[linesnumbered,ruled]{algorithm2e}

\theoremstyle{remark}

\theoremstyle{definition}
\newtheorem{definition}{Definition}
\theoremstyle{plain}
\newtheorem{theorem}{Theorem}

\newtheorem{prop}{Proposition}
\newtheorem{problem}{Problem}
\newtheorem{lemma}{Lemma}

\usepackage{dirtytalk}
\usepackage[hidelinks]{hyperref}
\newcommand{\todo}[1]{{\color{red}{#1}}}

\let\oldnl\nl
\newcommand{\nonl}{\renewcommand{\nl}{\let\nl\oldnl}}

\DeclareMathOperator*{\argmax}{argmax} 
\usepackage{mathtools}
\DeclarePairedDelimiter\abs{\lvert}{\rvert}%
\usepackage{bm}

\DeclarePairedDelimiterX{\norm}[1]{\lVert}{\rVert}{#1}

\usepackage{mathrsfs}
\usepackage[para,online,flushleft]{threeparttable}
\usepackage{array,booktabs,makecell}
\usepackage{diagbox}

\begin{document}

\title{\bf  Decision-Oriented Learning with Differentiable Submodular Maximization for Vehicle Routing Problem}

\author{Guangyao Shi,  Pratap Tokekar 
\thanks{This work is supported in part by National Science Foundation Grant No. 1943368 and Army Grant No. W911NF2120076.}
\thanks{Guangyao Shi,  Pratap Tokekar are with the University of Maryland, College Park, MD 20742 USA {\tt\small [gyshi}, {\tt\small tokekar]@umd.edu}}
}



\maketitle

\begin{abstract}
We study the problem of learning a function that maps context observations (input) to parameters of a submodular function (output). Our motivating case study is a specific type of vehicle routing problem, in which a team of Unmanned Ground Vehicles (UGVs) can serve as mobile charging stations to recharge a team of Unmanned Ground Vehicles (UAVs) that execute persistent monitoring tasks. {We want to learn the mapping from observations of UAV task routes and wind field to the parameters of a submodular objective function, which describes the distribution of landing positions of the UAVs .} Traditionally, such a learning problem is solved independently as a prediction phase without considering the downstream task optimization phase. However, the loss function used in prediction  may be misaligned with our final goal, i.e., a good routing decision. Good performance in the isolated prediction phase does not necessarily lead to good decisions in the downstream routing task. In this paper, we propose a framework that incorporates task optimization as a differentiable layer in the prediction phase. Our framework allows end-to-end training of the prediction model without using engineered intermediate loss that is targeted only at the prediction performance. In the proposed framework, task optimization (submodular maximization) is made differentiable by introducing stochastic perturbations into deterministic algorithms (i.e., stochastic smoothing). We demonstrate the efficacy of the proposed framework using synthetic data. Experimental results of the mobile charging station routing problem show that the proposed framework can result in better routing decisions, e.g. the average number of UAVs recharged increases, compared to the prediction-optimization separate approach.          
\end{abstract}

\section{Introduction}
Many multi-robot decision-making problems can be formulated as combinatorial optimization problems, among which the objectives
in some problems (e.g., mutual information \cite{krause2008near}, area explored,
number of targets tracked \cite{zhou2018resilient}, detection probability \cite{tseng2017near} etc.) have diminishing returns
property i.e., submodularity. Intuitively, submodularity formalizes the notion that adding more robots to
a larger multi-robot team cannot yield a smaller marginal gain
in the objective than adding the same robot to a smaller team.  

If the submodular objective is known and fixed, the multi-robot decision-making problem boils down to a submodular maximization problem, which is NP-hard but can be solved with an $(1-\frac{1}{e})$-approximation by the greedy algorithm \cite{nemhauser1978analysis}. However, in practice, there are several parameters that affect the objective function that may not be known exactly.  
\begin{figure}
    \centering
    \includegraphics[scale=0.8]{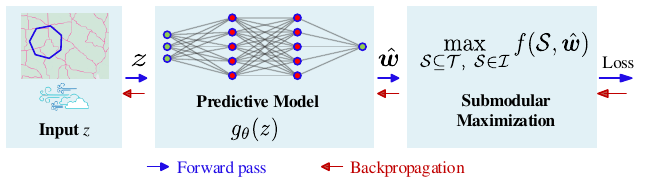}
    \caption{Decision-Oriented Learning framework. The training loss is defined after the downstream task.}
    \label{fig:outline}
\end{figure}
\begin{figure}
    \centering
    \includegraphics[scale=0.32]{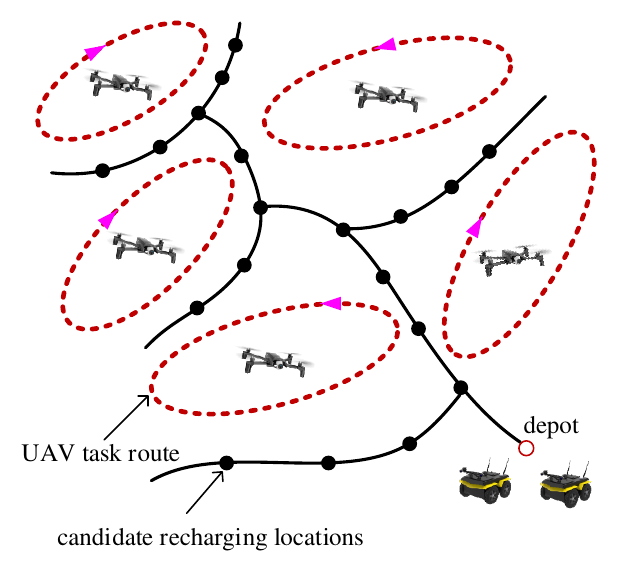}
    \caption{An illustrative example for vehicle routing problems. }
    \label{fig:motivation_UGV_routing}
\end{figure}

\begin{figure*}[ht]
    \centering
    \subfloat[Basis function $f_1$]{
    \includegraphics[width=0.23 \textwidth]{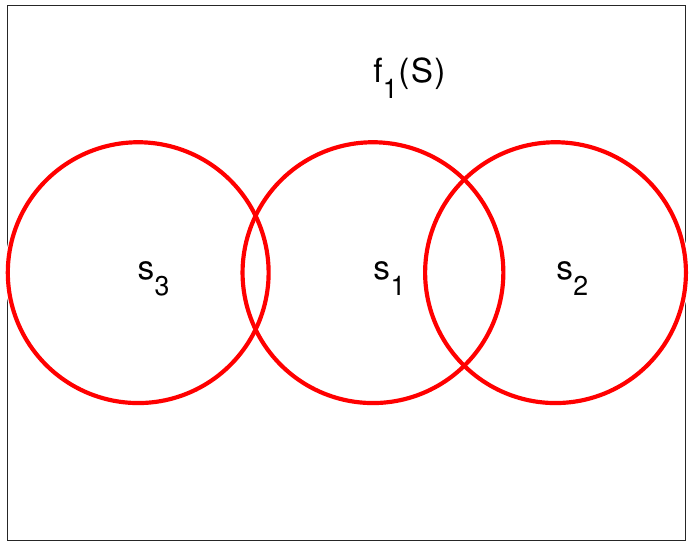}
    \label{fig:submodular_f1}
    }
    \subfloat[Basis function $f_2$]{
    \includegraphics[width=0.23 \textwidth]{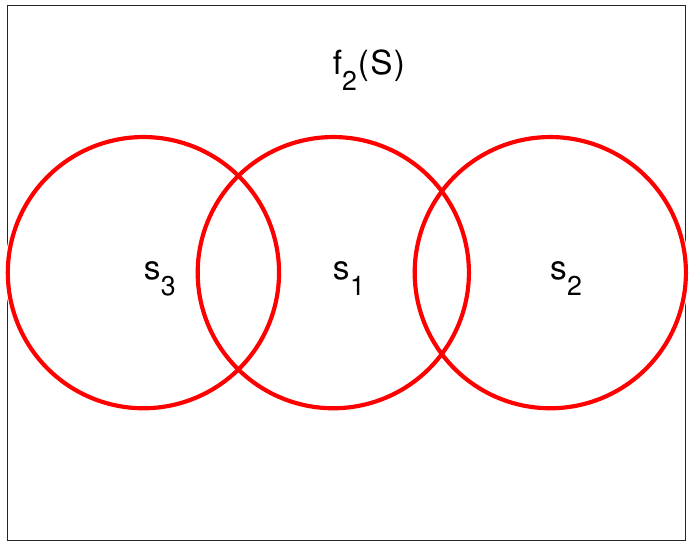}
    \label{fig:submodular_f2}
    } 
    \subfloat[Weighted sum of $f_1$ and $f_2$]{
    \centering
    \includegraphics[width=0.23 \textwidth]{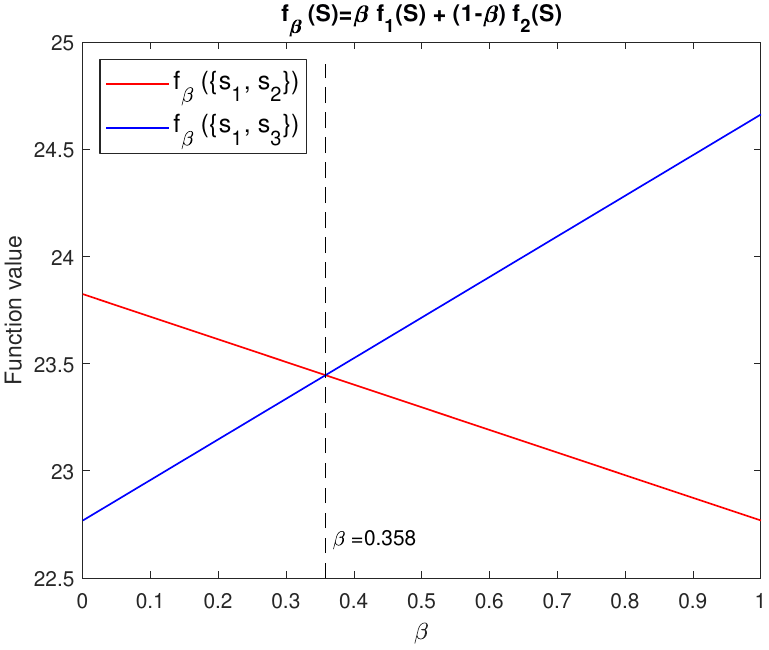}
    \label{fig:submodular_f_theta}
    } \\
    \subfloat[Ground truth data]{
    \centering
    \includegraphics[width=0.23 \textwidth]{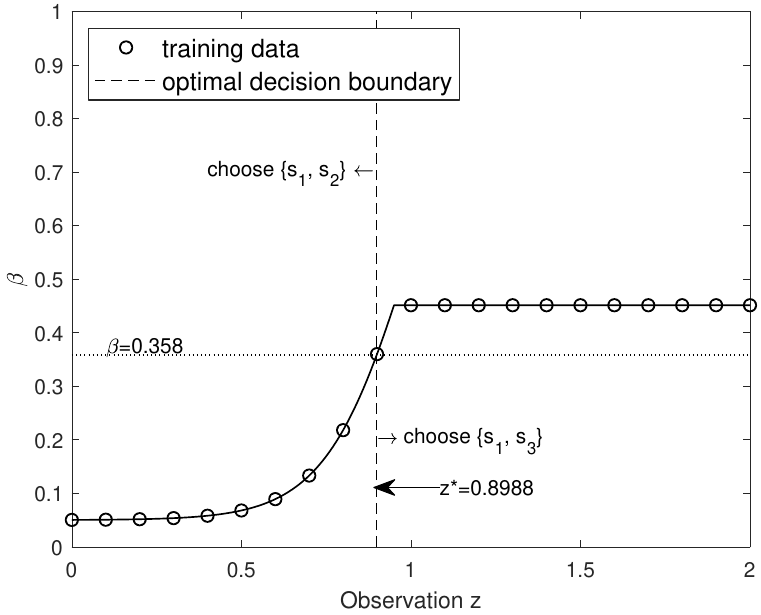}
    \label{fig:submodular_ground_truth}
    }
     \subfloat[Learned model using MSE]{
    \centering
    \includegraphics[width=0.23 \textwidth]{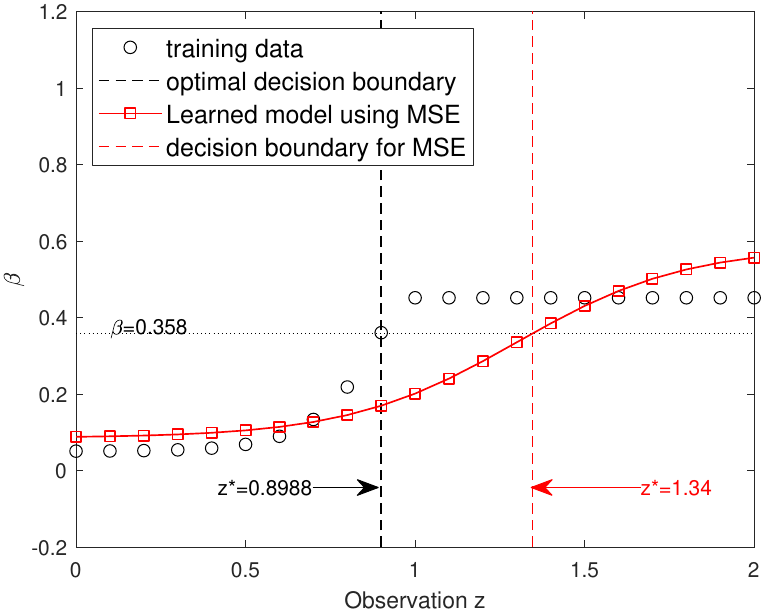}
    \label{fig:submodular_MSE}
    }
     \subfloat[Decision-Oriented-Learning results]{
    \centering
    \includegraphics[width=0.23 \textwidth]{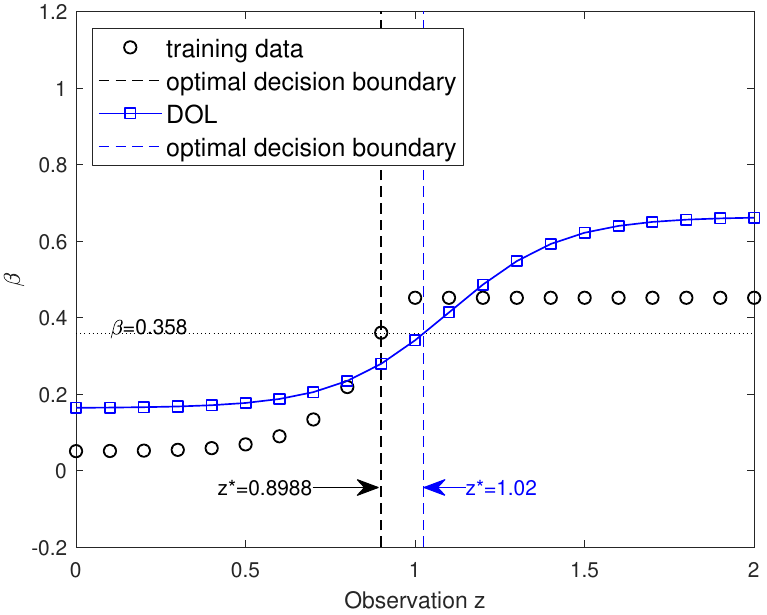}
    \label{fig:submodular_LDI}
    }
    \caption{
     An illustrative example to show the misalignment between the prediction model that achieves high predictive accuracy and the one that results in
good decisions. }
    \label{fig:misalignment_submodular_example}
    \vspace{-3mm}
\end{figure*}

Consider the following illustrative example of a vehicle routing problem shown in Figure~\ref{fig:motivation_UGV_routing}. Here, a team of Unmanned Ground Vehicles (UGVs) are tasked with servicing a set of requests that appear throughout the environment. We have a set of candidate routes of which we must select one for each UGV. The objective is to maximize the number of requests serviced. If a request location lies on more than one UGVs path (the paths may overlap as the UGVs move on a road network), it only counts once in the objective function. Thus, the objective function is 
a coverage function, which is a special case of the submodular function.

If we know the location of the requests, then we can solve this problem greedily to obtain a $(1-\frac{1}{e})$--approximation. The greedy algorithm requires the capability to compute the objective function $f(S)$. However, there are many scenarios where we may not know where the requests show up and as such not know $f(S)$. For example, the requests could correspond to Unmanned Aerial Vehicles (UAVs) that are carrying out persistent monitoring missions that land when out of charge so as to be recharged by mobile recharging stations~\cite{mitchell2015multi, derenick2011energy, liu2014energy, yu2019coverage, shi2022risk, asghar2022risk}. Here, even if we know the routes followed by each UAV, we may not know their exact landing locations since the energy consumption is stochastic~\cite{shi2022risk, asghar2022risk} and communication between UAVs and UGVs is not available (e.g., due to stealth). In such cases, we may be able to predict $f(S)$ using all the available information. We call the latter as \emph{context} $z$, which can include the routes of the UAVs, the environmental conditions including the wind conditions, etc. 

The traditional pipeline here would be to use the context information and predict $f(S)$ and then solve the downstream UGV route selection problem, $\argmax_{S} \hat{f}(S)$, using this predicted $\hat{f}(S)$. However, as the following example shows a good predictor of $f(S)$ does not necessarily align with making good decisions on the downstream task. On the other hand, a predictor that does not necessarily yield the best predictions of $f(S)$ may still yield the best decisions for the downstream $\argmax_{S} \hat{f}(S)$ problem.

We present an illustrative example of such misalignment in Fig. \ref{fig:misalignment_submodular_example}.  Let $f_1, f_2$ be two coverage functions ({coverage function is submodular by definition}) defined over set $\{s_1, s_2, s_3\}$. Given a subset $\mathcal{S} \subseteq \{s_1, s_2, s_3\}$, $f_{i}(\mathcal{S}), i=1, 2$ will return the area covered by the selection. The submodular objective that we are interested in is defined as $f_{\beta}(\mathcal{S}) = \beta f_{1}(\mathcal{S}) + (1-\beta)f_{2}(\mathcal{S}), \beta \in [0, 1]$, which is also submodular by definition.  Suppose that we want to maximize $f_{\beta}$ with a partition matroid: $\abs{\mathcal{S} \cap \{s_1\}} \leq 1$, $\abs{\mathcal{S} \cap \{s_2, s_3\}} \leq 1$. Then the optimal solution is either $\{s_1, s_2\}$ or $\{s_1, s_3\}$. In Fig. \ref{fig:submodular_f_theta}, we show how the optimal decision changes w.r.t. $\beta$. When $\beta \geq 0.358$, the optimal decision is $\{s_1, s_3\}$ since $f_{\beta}(\{s_1, s_3\}) \geq f_{\beta}(\{s_1, s_2\})$ (the blue line is above the red line). By contrast, when $\beta < 0.358$, the optimal decision is $\{s_1, s_2\}$. Next, let us look at the learning problem for $f_{\beta}$. We want to find a mapping from the observation $z$ to $\beta$. In Fig. \ref{fig:submodular_ground_truth}, we show the training data sampled from the ground truth and the optimal decision boundary $z^*=0.8988$, which is obtained by finding the intersection between the ground truth curve and $\beta=0.358$. 
If we use Mean Square Error (MSE) as the objective for learning without considering the downstream task, we will get two lines as shown in Fig. \ref{fig:submodular_MSE}. The decision boundary (dashed vertical red line, $z^*=1.34$, passing the intersection of the learned red line and $\beta=0.358$) is on the right of the optimal boundary, thus not optimal. By contrast, if we consider the downstream optimization, we will get two lines as shown in Fig. \ref{fig:submodular_LDI} and the decision boundary (dashed vertical blue line, $z^*=1.02$, passing the intersection of the learned blue line and $\beta=0.358$) is closer to the optimal boundary, thus reducing the regions of suboptimal decisions.
Such an observation motivates us to  incorporate the decision process (submodular maximization) into the learning process. 

To this end, we propose a Decision-Oriented-Learning (DOL) framework for learning context-aware parameterized submodular objectives. We focus on submodular functions that can be parameterized, i.e., $f(S,\bm{w})$, where the parameters are to be learned from the context. As described earlier and pointed out in~\cite{elmachtoub2022smart, wilder2019melding, mandi2020smart}, the best estimator of $\bm{w}$ does not necessarily yield the best decisions for the downstream task. Instead, in the proposed framework, the decision-making problem (submodular maximization) is treated as a differentiable layer that takes as input the output from the prediction module {as shown in Fig. \ref{fig:outline}}. The prediction module takes as input a context observation $z$ and predicts $\bm{w}$, i.e., the parameterized submodular function. By using a differentiable submodular optimization layer, we can train the prediction module using the loss from the downstream task, thereby yielding aligned predictions.

In summary, the main contribution is this paper is:
\begin{itemize}
    \item We propose a decision-oriented learning framework for mobile charging routing problems. We show how to formulate the learning problem for mobile charging routing and how to solve it.
    \item We demonstrate the effectiveness of our framework in several examples through simulation. 
\end{itemize}

The rest of the paper is organized as follows. We first
give a brief overview of the related work in Section  \ref{sec:related_work}. Then, we explain the problem setup and formulation in Section  \ref{sec:problem formulation}. We introduce the learning algorithm in Section \ref{sec:learning_algorithm} and validate the formulation and the proposed framework in Section \ref{sec:evaluation}.

\section{Related Work}\label{sec:related_work}

Most existing multi-robot decision-making work consider the case where the optimization objective is well-defined and known. Wilde et al. \cite{wilde2021learning} consider the case where the optimization objective is hard to quantitatively specify and may be subjective and proposed an interactive learning framework to learn the objectives. Our work shares a similar stance with \cite{wilde2021learning} but differs in two aspects. First, we consider the fact that the task objective may change in different contexts, for example in different weather conditions, and aim at learning a context-aware objective. Second, our learning framework integrates the downstream decision-making process into the learning process.   

Another line of research related to this work is decision-oriented learning. The key idea is to embed the decision-making problem as a differentiable layer in the learning pipelines. The main advantage is that it allows end-to-end training and reduces the engineering efforts to design some intermediate learning objectives. Such an idea was initially explored for continuous optimization problems \cite{amos2017optnet, agrawal2019differentiable} and has gained popularity in control and robotics \cite{muntwiler2022learning, amos2018differentiable, chen2018approximating, bhardwaj2020differentiable }. The idea was later extended to the combinatorial problems \cite{wilder2019melding, ferber2020mipaal, mandi2020smart, poganvcic2019differentiation}. Our work is inspired by \cite{wilder2019melding, ferber2020mipaal} and our framework integrates the decision-making process for mobile charging station routing, which is modeled as submodular maximization, into the learning process. 

This work is also closely related to differentiable submodular maximization. Submodular maximization and its variants have been
widely used in multi-robot decision-making problems including coverage, target tracking, exploration, and information gathering. 
These studies are all based on the fact that the greedy algorithm
and its variants can solve submodular maximization problems
its variants efficiently with a provable performance guarantee.
Since the submodular objective and greedy algorithm are tightly coupled, it is better to take into account the influence of the greedy algorithm when we consider learning submodular functions \cite{djolonga2017differentiable}. To this end, several differentiable versions of the greedy algorithms have been proposed \cite{djolonga2017differentiable, sakaue2021differentiable}. The core idea behind these algorithms is stochastic smoothing, i.e., perturb the algorithm by introducing some probability distribution in the intermediate steps.  Our framework is built on these differentiable greedy algorithms but is targeted specifically for context-dependent routing problems.

\section{Problem Formulation}\label{sec:problem formulation}
In this section, we first introduce the formulation of the learning problem. Then, we explain the setup of the case study and the parameterization of the objective.

We are interested in parameterized submodular objective function $f( \mathcal{S}, \bm{w})$, where $\bm{w}$ is the parameter vector. Readers are referred to \cite{wilde2021learning} for a formal definition. In practice, such an objective is usually unknown and context-dependent, i.e., the parameters $\bm{w} \in \mathcal{W}$ depend on the environment features. Our goal is to learn a function $g_{\bm{\theta}}: \mathcal{Z} \to \mathcal{W}$ that maps the context observation $z \in \mathcal{Z}$ to the objective parameters $\bm{w}$. Traditionally, finding the mapping $g_{\bm{\theta}}$ and optimizing the downstream objective $f(S, \bm{w})$ are considered separately: given the training data $\mathcal{D}=\{(\bm{z}_1, \bm{w}_1), (\bm{z}_2, \bm{w}_2), \ldots, (\bm{z}_{\abs{\mathcal{D}}}, \bm{w}_{\abs{\mathcal{D}}})\}$, first find the mapping  $g_{\bm{\theta}}$ by optimizing over $\bm{\theta}$ in a supervised fashion, and then use the parameter $\bm{w}=g_{\theta}(\bm{z})$ to optimize $f(S, \bm{w})$. 

By contrast, the proposed paradigm that integrates downstream optimization is given below. 

\begin{problem}\label{prob:main}
    Given the training data $\mathcal{D}=\{(\bm{z}_1, \bm{w}_1), (\bm{z}_2, \bm{w}_2), \ldots, (\bm{z}_{\abs{\mathcal{D}}}, \bm{w}_{\abs{\mathcal{D}}})\}$, learn a function $g_{\bm{\theta}}$ parameterized by $\bm{\theta}$ such that the learning cost $L=\frac{1}{\abs{\mathcal{D}}} \sum_{i=1}^{\abs{\mathcal{D}}} \ell_i(\bm{w}_i, \hat{\bm{w}}_i)$ is minimized, where $\ell_i(\bm{w}_i, \hat{\bm{w}}_i)$ is defined through Eq. \eqref{eq:mapping_theta} to Eq. \eqref{eq:learning_cost}:
    \begin{align}
    \hat{\bm{w}}_i & \coloneqq g_{\bm{\theta}}(\bm{z}_i) \label{eq:mapping_theta}\\
    \hat{\mathcal{S}} &\coloneqq \mathcal{S}^*(\hat{\bm{w}}_i) ~~\text{by solving \eqref{eq:SM}}~ \text{with}~ \bm{w}=\hat{\bm{w}}_i \label{eq:optimality_definition}\\
   \ell_i(\hat{\bm{w}}_i, \bm{w}_i) &\coloneqq f({\mathcal{S}^*(\bm{w}_i)}, {\bm{w}_i}) -f(\hat{\mathcal{S}}, {\bm{w}_i}), \label{eq:learning_cost}
    \end{align}
    where $\mathcal{S}^*(\bm{w}_i)$ denotes the solution of \eqref{eq:SM} returned by some approximation algorithms with $\bm{w}={\bm{w}}_i$; $f({\mathcal{S}^*(\bm{w}_i)}, {\bm{w}_i})$ denotes the decision quality when we use the ground truth parameter $\bm{w}_i$ for decisions; $f(\hat{\mathcal{S}}, {\bm{w}_i})$ denotes the decision quality when we use the predicted parameter $\hat{\bm{w}}_i$ for decisions, i.e., use $\hat{\bm{w}}_i$ to obtain the decision $\hat{\mathcal{S}}$, but the decision is evaluated w.r.t. the true parameter $\bm{w}_i$.
\end{problem}

 The intuition for Eq. \eqref{eq:learning_cost} is that we want to minimize the gap between the decision quality of the true parameters and that of the predicted parameters. One challenge is when we use the chain rule to compute the gradient of the loss function, we need to differentiate through the optimization problem (the first term on the r.h.s. of Eq. \eqref{eq:chain_rule_challenge}) as shown in the illustrative computational graph in Fig. \ref{fig:outline}.
 \begin{align}
    \frac{\partial \ell_i}{\partial \bm{\theta}} 
    = \frac{\partial \ell_i}{\partial \hat{\bm{w}}_i} \cdot \frac{\partial \hat{\bm{w}}_i}{ \partial \bm{\theta}} \label{eq:chain_rule_challenge}
\end{align}
 In the following sections, we will show how to approximately compute the first term on the r.h.s. of Eq. \eqref{eq:chain_rule_challenge}.

\section{Case Study}

 Suppose that there is a set of candidate routes, $\mathcal{T}$, each of which starts and terminates at the same depot. Our goal is to select a subset from $\mathcal{T}$ for UGVs to traverse and recharge the UAV along the way such that the total number of UAVs that UGVs will recharge is maximized.
 
\textbf{Environment Model:}  As shown in Fig. \ref{fig:motivation_UGV_routing}, the working environment is described by an area  $\mathcal{E} \subseteq \mathbb{R}^2$. There are $n_a$ UAVs that are executing persistent monitoring. The energy consumption of UAVs will be affected by the wind. The wind field is represented as a tuple $(\bm{\omega}_{s}, \omega_{o})$, where $\bm{\omega}_{s}$  and $\omega_{o}$ denote the description vectors for the speed and the orientation of the wind, respectively. 
There are $n_g$ UGVs in $\mathcal{E}$, denoted by the set 
$\{1, \ldots, n_g\}$. 

\textbf{UAV Behavior:}  There are three components defining the behavior of each UAV. The first one is the task route, which is defined as a sequence of ordered locations projected on the ground, and the UAV will persistently monitor these locations. The UAV will fly at a fixed speed $v_a$ between two task locations and its energy consumption will be affected by the wind. The second component is the recharging strategy dealing with the depletion of the battery. 
The third component is the energy consumption model. We use the same model as that in \cite{sorbelli2020energy}. 

\textbf{Context Observation:} Each observation $\bm{z}$ consists of two components: the task routes of all UAVs;  and the wind field $(\bm{\omega_{s}}, \omega_{o})$ of the working area.

If such submodular objective $f$ is known, the problem boils down to a submodular maximization problem with a matroid constraint: let 
$\mathcal{T} $ be set of all candidate routes, the problem is to select a subset from $\mathcal{T}$ to maximize the objective, i.e., 
\begin{equation}\label{eq:SM}
    \max_{\mathcal{S} \subseteq \mathcal{T}, ~\abs{\mathcal{S}} \leq n_g } f(\mathcal{S}, \bm{w}),
\end{equation}
where $\bm{w}$ denotes the parameters in the objective function. 

\textbf{Parameterization of the Objective Function}
In general, such applications have no closed-form expression of the objective function. In this paper, we consider the case where the objective function $f$ is the linear combination of a set of basis functions. Such parameterization techniques are commonly used in the literature on learning submodular functions \cite{tschiatschek2014learning, wilde2021learning, tseng2017near}.
Similar to \cite{wilde2021learning}, we assume without loss of generality that each basis function is characterized by a subset $W_i \subseteq V$. That is, for any $W_i$, let $\psi_i(\mathcal{S})$ be a count of how many vertices of the tours $\mathcal{S}$ lie in $W_i$, then $f_i$ is a functional of $\psi_i(\mathcal{S})$.  

The overall objective function is: 
\begin{equation}\label{eq:submodular_objective}
    f(\mathcal{S}, \bm{w}) = \sum_{i=1}^{n} \sum_{j=1}^{\abs{\Gamma}} w_{i, j} f_{i, j}(\mathcal{S}),
\end{equation}
where $\bm{w} = \left [ w_{i, j} \right ]$ denotes the matrix of unknown parameters of the function; basis functions are defined as
$    f_{i, j}(\mathcal{S}) = \sum_{\alpha=1}^{\psi_i(\mathcal{S})} \gamma_j^{(\alpha -1)}
$
; and $\gamma_j \in (0, 1]$ comes from a known set $\Gamma$.

\section{Learning Algorithm}\label{sec:learning_algorithm}
In this section, we describe the stochastic techniques to smoothe the greedy algorithm for submodular maximization and how can we apply the result to our framework. The key idea is: by introducing proper stochastic perturbances into combinatorial optimization,  the expected output as a function of its parameters can be smoothed and differentiable. 


\subsection{Smoothed Greedy Algorithm}
\begin{algorithm}[ht]\label{alg:smoothed_greedy}
    \caption{Smoothed Greedy}
    \SetKwInOut{Input}{Input}
    \SetKwInOut{Output}{Output}
    \Input{
   $f(\mathcal{S}, \bm{w})$ and independent set $\mathcal{I}$
    }
    \Output{Set $\mathcal{S}$ of tours for each robot}
    $\mathcal{S} \gets \emptyset$  \\
    \For{$k \gets 1$ \KwTo $N$}{
    \nonl // find all addable elements in the current round \\
    $U_k = \{u_1, \ldots, u_{n_k}\} \gets \{T \notin \mathcal{S} \mid \mathcal{S} \cup \{T\} \in \mathcal{I}\}$ \\
    \nonl // marginal gain for all addable elements \\
    $\bm{m}_k (\bm{w}) \gets (f_{\mathcal{S}}(u_1, \bm{w}), \ldots, f_{\mathcal{S}}(u_{n_k}, \bm{w}))$\\
    \nonl // compute a probability distribution \\
    $\bm{p}_k(\bm{w}) \gets \argmax_{\bm{p} \in \Delta^{n_k}} \{ \langle \bm{m}_k (\bm{w}),~\bm{p}  \rangle- \Omega_k(\bm{p}) \}$ \\
    $s_k \gets \text{sample}~ u \in U_k$~with probability $p_k(u, \bm{w})$ \\
    $\mathcal{S} \gets\mathcal{S} \cup \{s_k\}$
    }
    \textbf{return}~$\mathcal{S}$
\end{algorithm}
The Smoothed Greedy (SG) algorithm is given in Algorithm \ref{alg:smoothed_greedy}, which was first proposed in \cite{sakaue2021differentiable}.  For a given $\bm{w} \in \mathcal{W}$, In each iteration step, we compute marginal gain $f_{\mathcal{S}}(u, \bm{w})$ for each candidate element $u \in U_k$ (line 3); we define $n_k \coloneqq  \abs{U_k}$. Let $\bm{m}_k(\bm{w})=(m_k(u_1, \bm{w}), m_k(u_2, \bm{w}), \ldots, m_k(u_{n_k}, \bm{w})) \in \mathbb{R}^{n_k}$ denote the marginal gain vector. The probability vector, $\bm{p}_k(\bm{w})=(p_k(u_1, \bm{w}), \ldots, p_k(u_{n_k}, \bm{w}))$, is computed as: 
\begin{equation}\label{eq:step_k_prob}
    \bm{p}_k(\bm{w}) = \argmax_{\bm{p} \in \Delta^{n_k}} \{ \langle \bm{m}_k (\bm{w}),~\bm{p}  \rangle- \Omega_k(\bm{p}) \},
\end{equation}
where $\Delta^{n_k} \coloneqq \{\bm{p} \in \mathbb{R}^{n_k} \mid \bm{p} \geq \bm{0}_{n_k}, \langle \bm{p}, \bm{1}_{n_k} \rangle=1 \}$ is the $(n_k-1)$-dimensional probability simplex; $\Omega_k: \mathbb{R}^{n_k} \to \mathbb{R}$ is a strictly convex function and is a regularization function.  

Next, we will show the theoretical results for Algorithm \ref{alg:smoothed_greedy}. Detailed explanations and proofs can be found in \cite{sakaue2021differentiable}. Let $\delta \geq 0$ be a constant that satisfies $\delta \geq \Omega_k(\bm{p})-\Omega_k(\bm{q})$ for all $k=1, \ldots, \abs{\mathcal{S}}$ and $\bm{p}, \bm{q} \in \Delta^{n_k}$. We will use $\delta$ to quantify the performance of SG.

 As shown in Theorem 1 in \cite{sakaue2021differentiable}, in expectation, the output of SG satisfies that 
$ \mathbb{E}\left [ f(\mathcal{S}, \bm{w}) \right ] \geq (1-\frac{1}{e})f(OPT, \bm{w})-\delta n_g$,
 where $OPT$ denotes the optimal solution. This result 
 suggests that the SG algorithm in expectation almost preserves the performance of the deterministic greedy algorithm, whose approximation factor is $(1-\frac{1}{e})$, with one extra term $\delta n_g$, which is the price for differentiability. It should be noted that by using SG, the output is stochastic and we focus on the expected result of the output. 
The regularization functions $\Omega_k$ are chosen to guarantee the expected outputs of SG differentiable. Examples for $\Omega_k$ will be discussed in the Sec. \ref{sec:evaluation}.

\subsection{Gradient Estimation}
Let $\mathcal{O}_{\mathcal{I}}$ be the set of all possible solutions returned by SG.
Let $p(\mathcal{S}, \bm{w}) \in \left [ 0, 1\right ]$ be the probability for $\mathcal{S} \in \mathcal{O}_{\mathcal{I}}$. Specifically, for a returned sequence $\mathcal{S}=\{s_1, \ldots, s_{\abs{\mathcal{S}}}\} \in \mathcal{O}_{\mathcal{I}}$, the associated probability can be computed as 
$ p(\mathcal{S}, \bm{w}) = \prod_{k=1}^{\abs{\mathcal{S}}} p_k (s_k, \bm{w})$,
where $p_k (s_k, \bm{w})$ is the element of $\bm{p}_k(\bm{w})$ defined Eq. \eqref{eq:step_k_prob} corresponding to $s_k \in U_k$. 

Next, we will show how to construct a gradient estimator based on the output distribution. Let $Q(\mathcal{S})$ be any scalar- or vector-valued function. We want to compute $\nabla_{\bm{w}} \mathbb{E}_{\mathcal{S} \sim p(\bm{w})} \left [ Q(\mathcal{S}) \right ]=\sum_{\mathcal{S} \in \mathcal{O}_{\mathcal{I}}} Q(\mathcal{S}) \nabla_{\bm{w}} p(\mathcal{S}, \bm{w})$. Since the size of the independent set will increase exponentially w.r.t. the size of the ground set, it is computationally expensive to compute this gradient exactly. Instead, we will use the following unbiased estimator for the gradient in training.

As shown in Proposition 1 in \cite{sakaue2021differentiable},
 let $\mathcal{S}_j = (s_1, \ldots, s_{\abs{\mathcal{S}_j}}) \sim p(\bm{w}) (j=1, \ldots, N)$ be outputs of SG. Then, 
    \begin{equation}\label{eq:diff_SM_estimator}
        \frac{1}{N} \sum_{j=1}^N Q(\mathcal{S}_j) \otimes \nabla_{\bm{w}} \ln{p(\mathcal{S}_j, \bm{w})} 
    \end{equation}
    is an unbiased estimator of $\nabla_{\bm{w}} \mathbb{E}_{\mathcal{S} \sim p(\bm{w})} \left [ Q(\mathcal{S}) \right ]$, where $\otimes$ denotes the outer product.

\subsection{Differentiable Submodular Maximization for DOL}
 For the $i$-th training sample $(\bm{z}_i, \bm{w}_i)$, the associated cost w.r.t. $\bm{\theta}$ is redefined for SG as:
\begin{equation}\label{eq:expected_learning_cost_SM}
       \ell_i(\hat{\bm{w}}_i, \bm{w}_i) \coloneqq f({\mathcal{S}}^*(\bm{w}_i), \bm{w}_i) - \mathbb{E}_{\hat{\mathcal{S}} \sim p(\hat{\bm{w}}_i)} \left [ f(\hat{\mathcal{S}}, {\bm{w}_i}) \right ] ,
\end{equation}
where $\hat{\bm{w}}_i = g(\bm{z}_i, \bm{\theta})$.

For a training set with batch size $M$, we are interested in minimizing the empirical objective function $\frac{1}{M} \sum_{i=1}^M  \ell_i $, where $p(\hat{\bm{w}})$ is the output distribution of SG and $\ell_i$ is defined in Eq. \eqref{eq:expected_learning_cost_SM}. 

We compute the gradient using the chain rule: 
\begin{align}
    \frac{\partial \ell_i}{\partial \bm{\theta}} 
    = \frac{\partial \ell_i}{\partial \hat{\bm{w}}_i} \cdot \frac{\partial \hat{\bm{w}}_i}{ \partial \bm{\theta}}
    = -\frac{\partial \mathbb{E}_{\hat{\mathcal{S}} \sim p(\hat{\bm{w}}_i)} \left [ f(\hat{\mathcal{S}}, {\bm{w}_i}) \right ]  }{\partial \hat{\bm{w}}_i} 
 \cdot \frac{\partial \hat{\bm{w}}_i}{\partial \bm{\theta}}, \label{eq:chain_rule}
\end{align}
where $\hat{\bm{w}} = g(\bm{z}_i, \bm{\theta})$. 

Suppose we take $N$ trials of SG, by setting $Q(\hat{\mathcal{S}})=f(\hat{\mathcal{S}}, {\bm{w}_i})$ in Eq. \eqref{eq:diff_SM_estimator}, we have: 
\begin{equation}
   \begin{aligned}
     & \text{r.h.s.~of~} \eqref{eq:chain_rule}  \\&=  \frac{-1}{N} \sum_{j=1}^N f(\hat{\mathcal{S}}_j, \bm{w}_i) \otimes \nabla_{\hat{\bm{w}}_i} \ln{p(\hat{\mathcal{S}}_j, \hat{\bm{w}}_i)}  \cdot \frac{\partial \hat{\bm{w}}_i}{\partial \bm{\theta}},
   \end{aligned}
\end{equation}
where $\hat{\bm{w}}_i = g(\bm{z}_i, \bm{\theta})$. 

Then, the remaining problem is how to compute $\nabla_{\hat{\bm{w}}_i} \ln{p(\hat{\mathcal{S}}_j, \hat{\bm{w}}_i)}$ for a given sample $\hat{\mathcal{S}}_j = (s_1, \ldots, s_{\abs{\hat{\mathcal{S}}_j}})$ returned by SG.  It should be noticed that $
    \nabla_{\hat{\bm{w}}_i} \ln{p(\hat{\mathcal{S}}_j, \hat{\bm{w}}_i)}
    = \nabla_{\hat{\bm{w}}_i} \ln{\prod_{k=1}^{\abs{\hat{\mathcal{S}}_j}} p(s_k, \hat{\bm{w}}_i)} 
    =\sum_{k=1}^{\abs{\hat{\mathcal{S}}_j}} \frac{1}{p(s_k, \hat{\bm{w}}_i)} \nabla_{\hat{\bm{w}}_i} p(s_k, \hat{\bm{w}}_i),
$
where $p(s_k, \hat{\bm{w}}_i)$ can be obtained when we run the SG algorithm. Therefore, we just need to compute $\nabla_{\hat{\bm{w}}_i} p(s_k, \hat{\bm{w}}_i)$. Let $\bm{p}_k(\hat{\bm{w}}_i)$ be the probability returned by Eq. \eqref{eq:step_k_prob} at the step $k$ corresponding to the sample $\hat{\mathcal{S}}_j$. By using the chain rule,
\begin{equation}\label{eq:sample_prob_gradient}
    \nabla_{\hat{\bm{w}}_i} \bm{p}_k(\hat{\bm{w}}_i) = \nabla_{\bm{m}_k} \bm{p}_k(\bm{m}_k) \cdot  \nabla_{\hat{\bm{w}}_i} \bm{m}_k(\hat{\bm{w}}_i),
\end{equation}
 the row in $\nabla_{\hat{\bm{w}}_i} \bm{p}_k(\hat{\bm{w}}_i)$ corresponding to $s_k$ will give  $\nabla_{\hat{\bm{w}}_i} p(s_k, \hat{\bm{w}}_i)$. The first term on the r.h.s. of Eq. \eqref{eq:sample_prob_gradient} can be computed using auto-differentiation tools \cite{agrawal2019differentiable} (the objective in \eqref{eq:step_k_prob} is strictly concave) and the second term can be computed by differentiating the parameterized submodular objective. 
We can then use the above result to compute the gradient of a batch and update the parameters $\bm{\theta}$ using the stochastic gradient descent method. 

\section{Experiments} \label{sec:evaluation}

\subsubsection{Simulation Setup}~\\
 \noindent \textbf{Environment Model:} 
 There are $n_a =10$ UAVs and $n_g=3$ UGVs. The global wind $\omega$ is represented as $[a, b, \omega_o ]$, where $a$ and $b$ are the shape and scale parameters of Weibull distribution, respectively, and $\omega_o$ is the wind direction. 

\noindent \textbf{UAV and UGV Behavior:}  
In this case study, we consider the case that the task route is defined as a sequence of ordered locations uniformly sampled from a circle whose center is $[C_x, C_y]$ and radius is $r$. Using this geometric information, each route can be represented as a vector $C_x, C_y, r$. As for the recharging strategy,  we use a simple strategy in the simulation: whenever the state of charge drops below $30\%$, fly to the nearest recharging location and wait for the UGVs. We use the same energy model as that in \cite{sorbelli2020energy}. 
We generate UGV routes by first randomly selecting a set of nodes and then solving a Traveling Salesman Problem to get a route.

\noindent \textbf{Context Input $\bm{z}$ and mapping $g_{\bm{\theta}}(\bm{z})$:} As shown in Fig. \ref{fig:outline}, each $\bm{z}$ consists of two components: the task routes of all UAVs and the wind field vector $[a, b, \omega_{o}]$ of the working area. Since the route of the UAV can be parameterized by a circle $(C_x, C_y, r)$,   $z$ can be represented as $[C_x^1, C_y^1, r^1, C_x^2, C_y^2, r^2, \ldots, C_x^{n_a}, C_y^{n_a}, r^{n_a}, a, b, \omega_{o}]$. $g_{\bm{\theta}}(\bm{z})$ is instantiated using neural networks with one hidden layer (of size 64) with ReLu activation function. 

\noindent\textbf{Regularization and Basis Function} We choose the entropy function for experiments. Specifically, when $\Omega_k(\bm{p})=\epsilon \sum_{i=1}^{n_k}p(u_i)\ln p(u_i)$, where $p(u_i)$ is the $i-$th entry of $\bm{p} \in \left [ 0, 1 \right ]^{n_k}$ and $\epsilon > 0$ is an arbitrary constant, $\delta$ can be set to $\epsilon \ln n_k$. 
 For the road graph $G=(V, E)$, we use graph partition algorithms to generate nine sets of nodes, i.e., $\{W_1, \ldots, W_9\}$. For each partition $W_i \subset V$, we define three basis functions for decay parameters $\gamma \in \{0.001, 0.5, 1\}$.

\subsubsection{Generate Training Data}\label{sec:evaluation:generate training data}~\\
\noindent \textbf{Raw Data:} Based on the UAV UGV behavior models, we build a simulator for the routing problems. Given simulation parameters (e.g., UAV routes, and wind conditions), it will simulate UAVs' execution of persistent tasks. If we provide several selected routes to the simulator, it will return the number of UAVs recharged if UGV follows these routes. Based on this simulator, for each context observation $\bm{z}$, we test multiple possible route selections and obtain a set of the actual number of UAVs that UGV recharged. We will use this raw data to obtain the training data.

To train the decision-oriented framework proposed in Sec. \ref{sec:problem formulation}, we need the $i$-th data point to be in the form $(\bm{z}_i, \bm{w}_i)$, where $\bm{z}_i$ denotes the context input and $\bm{w}_i$ is the corresponding parameter vector of the objective function. However, $\bm{w}_i$ is not directly available and we need to do some pre-processing of the raw data. Specifically, for each $\bm{z}$, we have a set of values for different selections, i.e., $\mathcal{F} = \{f(\mathcal{S}_1), \ldots, f(\mathcal{S}_{\abs{\mathcal{F}}})\}$. We find the corresponding $\bm{w}$ by solving the following regularized least square optimization problem \cite{tschiatschek2014learning}, i.e.,
$
    \min_{\bm{w} \geq 0} \sum_{i=1}^{\abs{\mathcal{F}}} \norm{f(\mathcal{S}_i, \bm{w}) - f(\mathcal{S}_i)}^2_2 + \xi \norm{\bm{w}}^2_2, 
$
where $\xi$ is a user-specified regularization parameter. 

\begin{figure}
    \centering
    \includegraphics[scale=0.32]{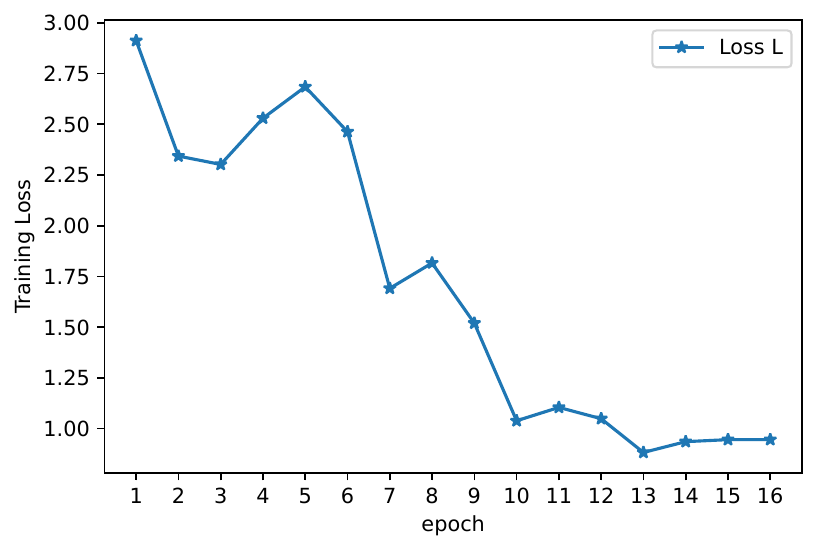}
    \caption{Per epoch loss curve of DOL. }
    \label{fig:smoothed_greedy_training_curve}
\end{figure}


\begin{table}[t]\label{table:baseline_comparison}
\centering
\caption{Test results for learned models. }
\begin{tabular}[t]{lccc}
\toprule
\small{Avg \# of UAV recharged}&DOL &two-stage &random \\
\midrule
{ $n_a=6, n_g=3$} &$5.3 \pm 0.5$ &$4.6 \pm 0.4$ &$2.1 \pm 1.4$ \\
\midrule
{ $n_a=10, n_g=3$} &$9.2 \pm 0.7$ &$8.5 \pm 0.6$ &$4.5 \pm 1.3$ \\
\midrule
{ $n_a=15, n_g=4$} &${14.1 \pm 0.8}$ &$13.2 \pm 0.7$ &$7.5 \pm 1.6$ \\
\bottomrule
\end{tabular}
\label{table:MC_convergence}
\end{table}

\subsubsection{Results}~

Fig. \ref{fig:smoothed_greedy_training_curve} shows the learning curves over epochs. In each epoch, we compute the gradient by sampling a batch size of 40 in each iteration. We can see that as the training epoch increases, the loss will gradually decrease to a steady value. It should be noticed that the loss here represents the solution quality gap between the solution obtained using ground truth parameters and the solution obtained using predicted parameters as defined in Problem \ref{prob:main}. Therefore, such a decrease suggests that the decision quality is improving. 

After training, we test the performance of the learned models using the simulator. We generate a set of context observations $\{\bm{z}_i\}_{\text{test}}$ and compute the corresponding predicted weights $\{\hat{\bm{w}}_i\}_{\text{test}}$. Then, we use $\{\hat{\bm{w}}_i\}_{\text{test}}$ to select routes and feed the route to the simulator to obtain the actual number of UAVs recharged. The result is shown in Table \ref{table:baseline_comparison}. We compare three approaches: DOL (our), two-stage (classic supervise learning with MSE loss), and random (select routes randomly without any learning ).  As shown in Table \ref{table:baseline_comparison}, our approach on average can result in better route selection and recharge more UAVs.  

\section{Conclusion}
We propose a decision-oriented learning framework for
mobile charging routing problems. We first show how to
formulate the learning problem in the context of mobile charging routing. Then, we show how to make submodular maximization a differentiable layer by using stochastic smoothing techniques. The proposed framework and formulation are validated through several case studies. 

\bibliographystyle{IEEEtran}
\bibliography{IEEEabrv, main}

\newpage
\end{document}